**A multi-scale mapping approach based on a deep learning CNN model for reconstructing high-resolution urban DEMs**

Ling Jiang [a,b,c], Yang Hu [d], Xilin Xia [c], Qiuhua Liang [a,c,*], Andrea Soltoggio [d]

[a] *State Key Laboratory of Hydrology-Water Resources and Hydraulic Engineering, Hohai University, Nanjing, 210098, China*

[b] *Anhui Engineering Laboratory of Geo-information Smart Sensing and Services, Chuzhou University, Chuzhou, 239000, China*

[c] *School of Architecture, Building and Civil Engineering, Loughborough University, Leicestershire, UK*

[d] *School of Computer Science, Loughborough University, Leicestershire, UK*

**\*** *Correspondence: q.liang@lboro.ac.uk (Q.L. Liang)*

**Abstract**: The shortage of high-resolution urban digital elevation model (DEM) datasets has been a challenge for modelling urban flood and managing its risk. A solution is to develop effective approaches to reconstruct high-resolution DEMs from their low-resolution equivalents that are more widely available. However, the current high-resolution DEM reconstruction approaches mainly focus on natural topography. Few attempts have been made for urban topography which is typically an integration of complex man-made and natural features. This study proposes a novel multi-scale mapping approach based on convolutional neural network (CNN) to deal with the complex characteristics of urban topography and reconstruct high-resolution urban DEMs. The proposed multi-scale CNN model is firstly trained using urban DEMs that contain topographic features at different resolutions, and then used to reconstruct the urban DEM at a specified (high) resolution from a low-resolution equivalent. A two-level accuracy assessment approach is also designed to evaluate the performance of the proposed urban DEM reconstruction method, in terms of numerical accuracy and morphological accuracy. The proposed DEM reconstruction approach is applied to a 121 km$^2$ urbanized area in London, UK. Compared with other commonly used methods, the current CNN based approach produces superior results, providing a cost-effective innovative method to acquire high-resolution DEMs in other data-scarce environments.

**1. Introduction**

Digital elevation models (DEMs) have been widely used in many fields such as landform evolution, soil erosion modeling and other geo-simulations (Bishop et al., 2012; Liu et al., 2015; Mondal et al., 2017; Li and Wong, 2010). In particular, DEMs provide indispensable data to support water resources management and flood risk assessment (Moore et al., 1991; O'Loughlin et al., 2016). In urban flood risk assessment, the availability of high-resolution urban DEMs is crucial for the accurate representation of complex urban topographic features and required for a reliable prediction of flood inundation to inform risk calculation (Ramirez et al., 2016; Leitão and de Sousa, 2018).

The common ways of acquiring high-resolution urban DEMs include ground surveying and remote sensing through light detection and ranging (LiDAR), interferometric synthetic aperture radar (InSAR) and other techniques (Shan and Aparajithan, 2005; Rossi and Gernhardt, 2013; Zhu et al., 2018; Bagheri et al., 2018; Le Besnerais et al., 2008). These approaches are usually labor-intensive and financially expensive, hindering their wider application at a large scale (e.g., across an entire city). As such, high-resolution urban DEMs are not always available, especially for cities in the developing countries. This essentially imposes a barrier for many applications including the development of effective urban flood risk management strategies that are necessary to be informed by high-resolution flood modelling results. Hence, it is necessary to develop alternative and more cost-effective approaches to construct high-resolution urban DEMs to support a wide range of applications.

Although high-resolution urban DEMs are not always available, low-resolution DEMs, on the other hand, are relatively easy to access. For example, there exist a range of open-access global or regional DEMs, including Shuttle Radar Topography Mission (SRTM), ALOS World 3D and pan-Arctic DEM (Hawker et al., 2018). Thus, it may be desirable to develop effective techniques to enhance the quality of low-resolution DEMs to subsequently obtain high-resolution urban DEMs. Most of the existing high-resolution DEM reconstruction methods are developed for natural terrains, which may be generally classified into three categories, i.e., DEM interpolation, DEM enhancement and learning-based DEM reconstruction.

The DEM interpolation methods, commonly including inverse distance weighting (IDW), bilinear interpolation (BI) and cubic convolution (CC), are generally implemented according to spatial autocorrelation, i.e., the correlation of the ground elevations between two points is inverse to the distance between them (also known as Tobler's first law of geography) (Aguilar et al., 2005; Heritage et al., 2009; Wise, 2011; Arun, 2013; Tan et al., 2018). DEM interpolation methods have been widely applied to generate high-resolution DEMs, but the resulting products cannot contain fine details that the low-resolution DEMs lose to accurately reflect the real elevations. DEM enhancement methods attempt to restore the lost topographic features via introducing extra information to enhance the quality of low-resolution DEMs. The extra information may be derived from additional elevation points, contours, land-use maps and flood extents (Tran et al., 2014; Yue et al., 2015; Mason et al., 2016; Yue et al., 2017; Li et al., 2017), etc. DEM enhancement methods can effectively reconstruct high-resolution DEMs by fusing multiple DEMs and datasets at different scales or from various sources. Nevertheless, the required extra high-accuracy topographic information for implementation of this type of methods is still hard to acquire, especially for a large extent. The learning-based approaches generate high-resolution DEMs by establishing the correlation between low- and high-resolution DEMs through a training process (Xu et al., 2015; Chen et al., 2016; Moon and Choi, 2016; Liu et al., 2017). Learning-based models can be trained to learn from multi-dimensional information, which may potentially produce high-resolution DEMs of better quality than the aforementioned alternative approaches. However, less research has been done in this topic, and the existed learning-based models are relatively simple and not suitable for applications in complex urban environments.

Most of the existing DEM reconstruction methods are developed and applied to natural terrains. Reconstruction of urban high-resolution DEMs faces extra challenges, and direct application of the existing methods in the complex urban environments is questionable and may not be feasible. Due to human interventions, urban topography is typically an intricate synthesis of man-made and natural features. In most of the cases, man-made features are more predominant, which may create abrupt changes to the topography at different scales. For flood modelling, the key urban structures/features may pose particular influence on and even control the underlying hydrological processes and must be accurately represented in urban DEMs (Mark et al., 2004; Ozdemir et al., 2013; Leitão and de Sousa, 2018). Therefore, there is a strong need to develop new approaches to support multi-scale reconstruction to efficiently reconstruct urban DEMs at specified resolutions from a low-resolution equivalent.

Whilst cities are covered by man-made topographic features of different types and scales, they are planned and built according to specific regulations and codes, and urban topography commonly presents a high level of self-similar features, especially for cities in the same region. This is particularly suitable for the application of learning-based approaches. For example, Convolutional Neural Network (CNN) (LeCun et al., 2015; Schmidhuber, 2015) is a deep learning technique designed to automatically and adaptively learn spatial hierarchies of image features and has been successfully applied in image recognition and many other fields, such as machine translation and autonomous driving (Abdel-Hamid et al., 2014; Chen et al., 2015; Gu et al., 2018). An urban gridded DEM can be effectively regarded as an image. With the availability of localized DEMs of different resolution, a CNN model may be trained to recognize the urban topographic features and used to reconstruct high-resolution DEMs from low-resolution ones across a much larger area. Moreover, it is feasible to acquire high-resolution urban DEMs in localized (small) areas by diverse survey techniques nowadays. Therefore, this paper aims to develop an innovative approach by combining a deep-learning CNN model and localized high-resolution urban DEMs to substantially improve the quality of low-resolution urban DEMs (i.e., the main focus is resolution enhancement, not noise filtering) and subsequently reconstruct high-resolution urban DEMs for a large area. This is first approach to employ CNN to enhance the accuracy of low-resolution urban DEMs, and can greatly promote the in-depth exploration to the levels of urban planning and management (e.g. urban flood risk management).

The rest of this paper is arranged as follows: Section 2 introduces the proposed multi-scale mapping approach for urban DEM reconstruction, followed by the introduction of two-level accuracy assessment framework in Section 3; Section 4 describes the experiments undertaken to assess the effectiveness of high-resolution urban DEM reconstruction; further discussion is given in Section 5; and finally several remarks are summarized in Section 6.

## 2. A CNN-based multi-scale mapping approach

A Multi-Scale Mapping approach based on CNN (MSM-CNN) is developed to reconstruct high-accuracy urban DEMs at a specified higher resolution from a low-resolution dataset, which is illustrated in Fig. 1. Herein, the low-resolution urban DEM is denoted as *X*, and the corresponding datasets at higher resolutions are denoted as $Y^2$, $Y^4$, ... , $Y^{2^n}$, where the superscript $2^n$ indicates that the urban DEMs are at $2^n$ times higher resolution than DEM *X* and *n* is a positive integer. The goal here is to reconstruct any high-resolution urban DEM $F^{2^n}(X)$ from the low-resolution DEM *X* to ensure $F^{2^n}(X)$ is as close to the ground truth dataset $Y^{2^n}$ as possible, which will be achieved by training a CNN to learn mapping *F*.

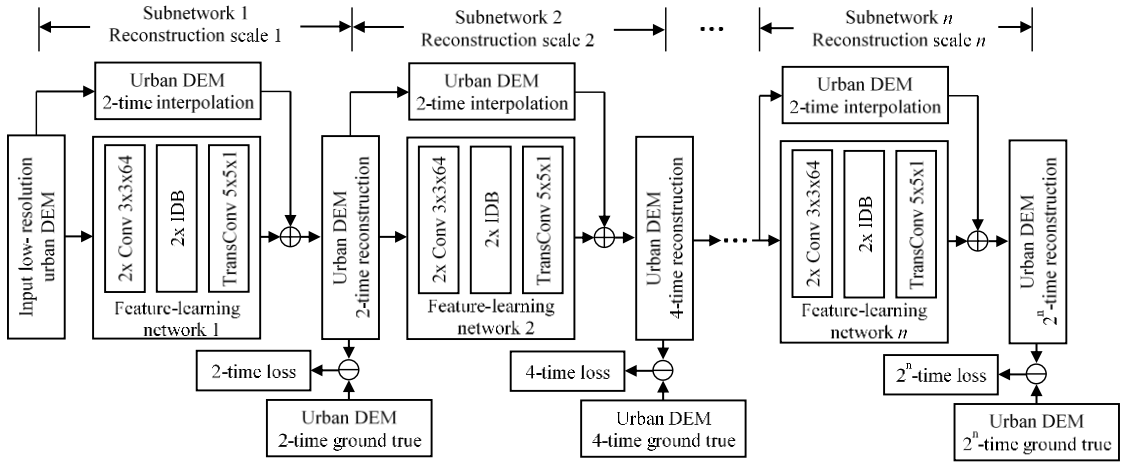

(a) Multi-scale gradual network

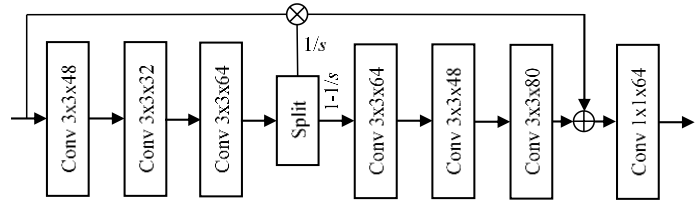

(b) Information distillation block (IDB)

**Fig. 1.** Flowchart of the MSM-CNN. The symbols, ⊕, ⊖ and ⊗, represent the element-wise sum operator, the loss-calculation operator and the concatenation operator, respectively. The abbreviations of Conv and TransConv denote the convolutional layer and transposed convolution layer. The expression 2x Conv or 2x

IDB represents two convolutional layers or two IDBs. The size of convolutional layers or transposed convolution layers is in the format of width by height by number of filters, such as 3 x 3 x 64. The symbol *s* stands for the number of parts into which the feature maps are split.

*2.1. Network architecture*

The detailed network architecture is shown in Fig. 1a, which consists of several subnetworks. Each of these subnetworks performs a 2-time reconstruction to its input urban DEM. According to the existing state-of-the-art results, a network with skip connections bypassing certain intermediate layers could lead to better performance (Krizhevsky et al., 2012; Simonyan and Zisserman et al., 2015; He et al., 2016). Therefore, skip connections are also introduced between the input and output of each of the subnetworks. Specifically, the input urban DEM of each subnetwork is interpolated to become 2 times of its original resolution using a nearest neighbor (NN) method, and the interpolated urban DEM is then directly summed to the output of the feature-learning network. The skip connections encourage the feature-learning networks to effectively learn and predict the missing topographic details from the low-resolution urban DEMs to generate high-resolution datasets. Note that NN here is chosen due to its computational efficiency compared to other interpolation methods. Since each subnetwork only performs a 2-time reconstruction, the proposed architecture can effectively train a single network to construct urban DEMs at different higher resolutions.

In the proposed architecture, the feature-learning network is a key component in each of the subnetworks. Each feature-learning network starts with 2 convolutional layers with the kernel size specified as in Fig. 1a. The effect of the two convolutional layers is to extract initial features for further feature learning. The first two convolutional layers in the feature-learning network is followed by two information distillation blocks (IDBs) (Hui et al., 2018) to learn more powerful deep features for urban DEM reconstruction. The architecture of IDB is presented in Fig. 1b. The IDB starts with a stack of six convolutional layers, with the filter size specified as in Fig. 1b. After the first three layers in each IDB, the output feature maps are split into two parts. The $1 - 1/s$ percent of the feature channels are used as the input to the next three layers, while the other $1/s$ percent feature channels is directly concatenated with the output of the next three layers. Such a structure creates skip connections and combines features in both shallower and deeper layers. The output of the first six blocks in IDB is passed to a seventh convolutional layer. This convolutional layer with 1 x 1 filters acts similarly to a bottleneck layer (Szegedy et al., 2015); its effect is to combine and compress the shallow and deep features output by the previous layers. Herein, although we use IDB as the backbone of the proposed network, other architectures could potentially also be used to replace IDB for feature learning. Note that this paper focuses on a multi-scale network for urban DEM reconstruction rather than seeking the backbone architecture with the best performance; we select IDB due to its reported excellent performance in accuracy and efficiency in computational cost. After the two IDBs, a transposed convolutional layer is applied to project the output feature maps of a subnetwork to a reconstruction at 2-time resolution with respect to the input of this subnetwork.

The proposed network introduces non-linearity using rectified linear unit (ReLU) formulated as $y = max(0, x)$ where $x$ represent the input feature maps and $y$ the output. $y$ equals to $x$ if $x$ is positive, otherwise $y$ is 0. ReLU is adopted due to its widely reported effectiveness in literatures (Krizhevsky et al., 2012; Simonyan and Zisserman, 2015; He et al., 2016). Herein, all of the convolutional layers are followed by a ReLU unless it is specifically mentioned.

An advantage of the proposed multiple-scale architecture with respect to single-scale architectures is that the multi-scale supervision has been introduced to regularize the intermediate features of the urban DEM, which can faithfully enhance the output of each subnetwork to become as close to the high-resolution 'true' DEM as possible. The adopted multi-scale supervision enables effortless and effective reconstruction of quality-enhanced urban DEMs at any specified high resolution. Note that multiscale design and computing losses at intermediate network layers to guide the learning process have been widely used in deep neural network architectures (Szegedy et al., 2015; Lee te al., 2015; Xie et al., 2015; Lai et al., 2017). In this paper, for the first time, we introduce this principle to the domain of urban DEM reconstruction.

*2.2. Loss function*

This section introduces the loss function used to train the network. Let $R_i$ be the $2^i$-time reconstruction result and $Y_i$ be the corresponding ground truth. The loss denoted by $L_i$ between $R_i$ and $Y_i$ is calculated as follows:

$$L_i = \frac{1}{N} \sum_{j=1}^{N} |R_{i,j} - Y_{i,j}| \qquad (1)$$

where $R_{i,j}$ and $Y_{i,j}$ are the element in $R_i$ and $Y_i$, respectively, and $N$ is the cell number. The overall loss $L$ is the sum of the losses at all scales:

$$L = \sum_{i=1}^{n} L_i \qquad (2)$$

Theoretically, a weighted sum could achieve better balance among the losses at different scales. However, preliminary experiments revealed that the sum loss with equal weights is sufficient to achieve a good performance. The loss function is based on mean of absolute error (MAE).

We note that other loss functions like structure similarity index (SSIM) and peak signal-to-noise ratio (PSNR) could potentially be used. However, this paper focuses on a multi-scale network architecture for urban DEM reconstruction instead of seeking the optimal loss function.

*2.3. Network training*

We train all the layers in the proposed network from scratch based on standard back-propagation with Adam optimizer (Kingma and Adam, 2015). The weights for convolutional layers are initialized using the method in He et al. (2015). The network training is conducted with a batch size of 64. The weight decay is set to 0.0001, and the learning rate is set to 0.0001 initially and reduced by a factor of 10 after 250 thousand iterations.

In the training process, we divide the training dataset of urban DEMs (see subsection 4.1) into blocks with a size of 500 by 500. Each block also has an overlapping of 250 cells in both horizontal and vertical direction with its neighbors. During each forward-backward pass of the network, a batch of 64 blocks is randomly selected for each training area, and then a patch from each block is randomly cropped, followed by forming the batch of training data through concatenating all of the patches. The size of a patch is chosen to meet the computational capacity, which depends on the number of scales in the network.

**3. Two-level accuracy assessment**

To evaluate the performance of the proposed urban DEM reconstruction method, a two-level assessment approach is designed to quantify the numerical accuracy and morphological accuracy of the resulting products. Herein, the numerical accuracy is a quantification of elevation error at cell locations, while the morphological accuracy is a region-scale quantification of morphology variance between the urban DEM and ground truth.

*3.1. Numerical accuracy*

Numerical accuracy is assessed by the difference of pointwise elevation between the reconstructed and 'true' urban DEMs. Three well-known metrics, i.e., mean absolute error (MAE), root mean square error (RMSE) and standard deviation (STD), are employed to quantify the numerical accuracy. The related equations are given as follows:

$$MAE = \sum_{i=1}^{n}|x_i - y_i|/n \tag{3}$$

$$RMSE = \sqrt{\frac{1}{n}\sum_{i=1}^{n}(x_i - y_i)^2} \tag{4}$$

$$STD = \sqrt{\frac{1}{n}\sum_{i=1}^{n}\left[(x_i - y_i) - \overline{(x - y)}\right]} \tag{5}$$

where *n* is the total count of valid grid cells, *x* denotes the elevation of the reconstructed urban DEM and *y* refers to the reference data.

*3.2. Morphological accuracy*

A DEM not only represents the ground elevation at each of its cells, but also reveals the structure of the topography. As the skeleton of topography, topographic structure decides the spatial pattern of geomorphology (Wilson, 2012). Hence, the accuracy in representing the topographic structure is an essential indicator for DEM quality assessment. In the case of urban topography, the topographic structure may be mainly reflected by road networks and building clusters that have a significant impact on surface runoff and flow processes. Accordingly, the morphological accuracy, i.e., the assessment of topographic feature difference, can be evaluated by measuring the variances of the road profiles and building boundaries derived from the reconstructed urban DEM and the reference data.

The road-profile variance is measured through the following steps: 1) delineate the road centerlines; 2) densify vertices along each road centerline stepped by the cell size of the reconstructed urban DEM; 3) obtain the road centerline profiles from both the reconstructed DEM and the reference data; and 4) apply the Pearson's correlation coefficient (PCC) to quantify the variance between profiles for each road, and take the average and standard deviation (STD) of PCCs to denote the difference. The PCC is calculated as follows:

$$PCC = \frac{\sum_{i=1}^{m}(x_i - \bar{x})(y_i - \bar{y})}{\sqrt{\sum_{i=1}^{m}(x_i - \bar{x})^2}\sqrt{\sum_{i=1}^{m}(y_i - \bar{y})^2}} \qquad (6)$$

where $m$ represents the number of the profile vertices, $x$ and $y$ are the values corresponding to the reconstructed and reference profiles being compared.

The variance of the building boundaries can be measured through three steps:

Step 1 is to count the reference data by: 1) preprocessing the building polygons by merging the adjacent polygons and deleting those small and discrete patches according to an area threshold of 20 m$^2$; 2) obtaining the boundary line of each building patch and converting all boundary lines to a raster format using the cell size of the reconstructed DEM; and 3) counting the boundary grid cells as the reference truth.

Step 2 is to extract the building boundaries from the reconstructed urban DEM by: 1) highlighting the boundaries between features (e.g., the boundary where a building meets a road) by an edge-enhancement (or high-pass) filter in the ArcGIS software; 2) screening the candidates of boundary grid cells via an edge threshold of 1; and 3) obtaining the boundary cells using a thinning tool available in the ArcGIS software.

Step 3 is to quantify the variance by: 1) selecting the boundary cells from Step 2 according to the location of the reference boundary lines with no buffer, and buffers of 1, 2, and 3 times of the cell size of the reconstructed DEM, respectively; and 2) calculating the ratio between the number of selected cells and that of the reference truth successively.

## 4. Experiments and results

In order to validate the performance of the proposed MSM-CNN method, a series of simulation experiments have been undertaken. In the experiments, the MSM-CNN model is trained and applied to reconstruct high-resolution urban DEMs in the selected case study area. The experiments are performed on a single GPU server with Nvidia K80 GPUs.

The produced outputs are compared with the results from several other popular interpolation or resample methods, including IDW, BI and CC. Herein, the decision that some other geostatistical interpolations (e.g. Kriging) are not chosen is attributed to their poor accuracy and extreme time consumption for the study area according to the results of preliminary experiments. The experimental setup is illustrated in the flowchart as shown in Fig. 2. In the experiments, urban DEMs at low resolutions of 2, 4 and 8 m are used to reconstruct high-resolution urban DEMs of 0.5 m to evaluate the performance of the multi-scale gradual network. It should be pointed out that, due to the lack of real datasets of 2, 4 and 8 m in the same period, in order to ensure the consistency of evaluation benchmark, the reconstructions from 8 to 0.5 m, 4 to 0.5 m and 2 to 0.5 m are designed to demonstrate the performances of the proposed approach. Herein, in the reconstructing phase, the test dataset is divided into blocks with a size of 250 by 250 with an overlap of 125 cells against their neighbors, and finally, each block is constructed individually and combined together to obtain the reconstructed urban DEM.

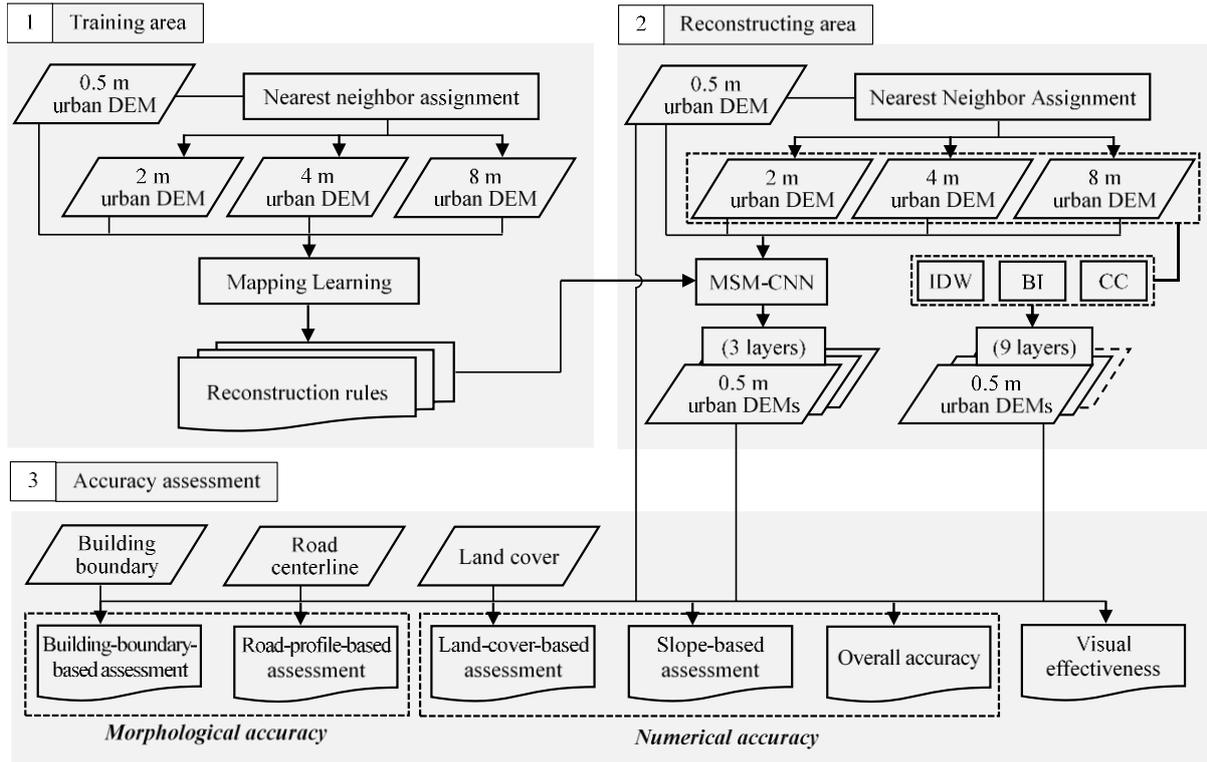

**Fig. 2**. Experimental setup.

*4.1. Study area and data*

As one of the largest cities in the world, London, UK is highly urbanized with a population of 8 million and is selected as the study area. We firstly train the MSM-CNN model using three small areas in the city. The three chosen training areas with significant differences in topographical features are located in the suburban, urban and rural regions, respectively. Each training site covers a 5 by 5 km area. After being trained, the MSM-CNN model is applied to reconstruct high-resolution DEMs in another larger area of 121 km$^2$, which is an urbanized area with mixed topographic features. The rationale to perform training and test in different areas is that, although the overall urban designs could vary in different areas, the local features like lines, edges and blocks are similar across different natural and manmade structures; since a CNN focuses on local features, it could be used to reconstruct the urban structures in an area that is unseen in the training data. In this reconstructed area, 8 samples of 1 by 1 km blocks are selected to facilitate morphological accuracy assessment. Fig. 3 shows the locations of the training, reconstruction and sample areas in the City of London.

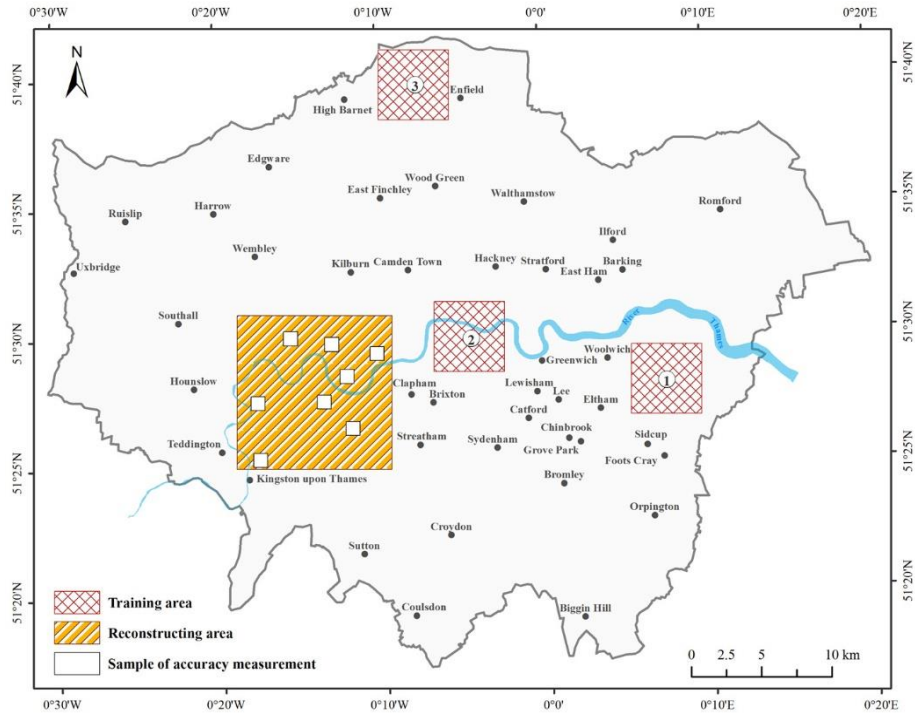

**Fig. 3.** Location of study area.

In this work, a 0.5 m LiDAR DSM is used as the baseline high-resolution urban DEM, which is published by the Environment Agency, UK (https://environment.data.gov.uk/ds/survey/index.jsp#/survey). This dataset is employed for training the MSM-CNN model and used as the reference truth for assessing the reconstruction accuracy. The low-resolution DEMs for training and testing the MSM-CNN model are also obtained from this 0.5 m DEM by resampling it to 2, 4 and 8 m resolutions using the NN downsampling (Fig. 4). We select NN instead of other alternatives like BI and CC for the following reason. This paper focuses on urban DEM which includes lot of abrupt elevation changes (e.g. a road with high buildings at both sides). For this specific type of data, methods like BI and CC could be less suitable compared to NN, as they introduce 'fake' elevation for the areas with abrupt features. Relevant datasets of land cover, road centerline and building are downloaded from Digimap (https://digimap.edina.ac.uk) for use in the current study.

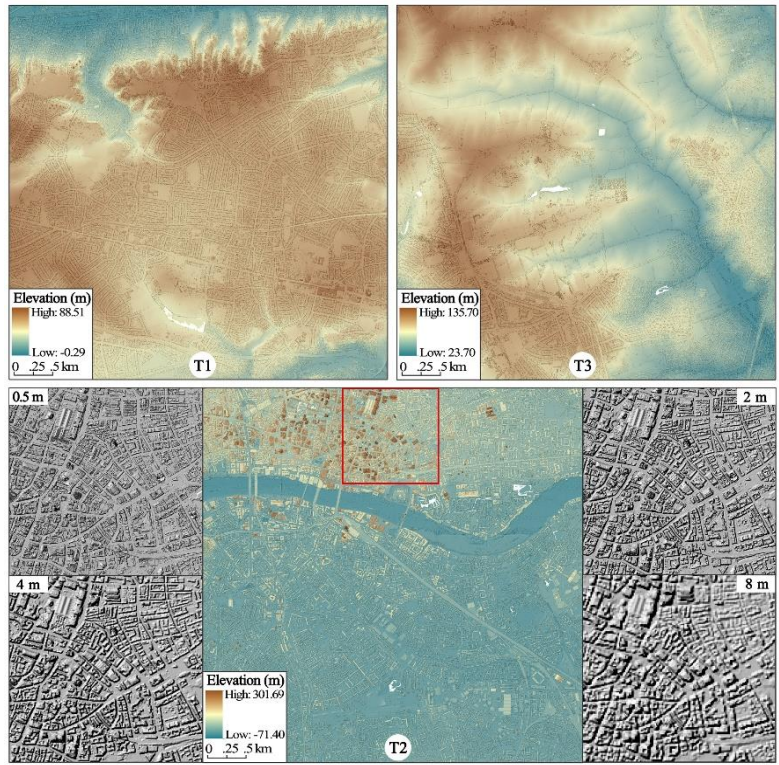

**Fig.4.** Urban DEMs of training areas, where T1, T2 and T3 denote the training area 1, 2 and 3, respectively.

*4.2. Visual assessment*

The reconstructed 0.5 m urban DEMs using different methods are plotted together with the low-resolution urban DEMs of 8, 4 and 2 m in Fig. 5. Obviously, the detailed features of urban topography are gradually lost as the resolution of the DEMs reduces from 0.5 to 2, 4 and 8 m (Fig. 5i, e and a). The topographic structure related to road network and building group becomes blurry when the DEM resolution decreases. On the 8 m urban DEM, the roads and buildings have become hard to identify. As depicted in Fig. 5k-l, g-h and c-d, the use of BI, CC and IDW interpolation methods shows certain level of enhancement in the topographic details. However, the enhancement is generally very limited. Particularly, it is not possible to restore the topographic structure from the lowest resolution (8 m) urban DEM. Moreover, the reconstruction result of IDW reproduces the hillocks and the BI and CC results exhibit with net-like features, which do not conform to the morphological cognition of urban topography.

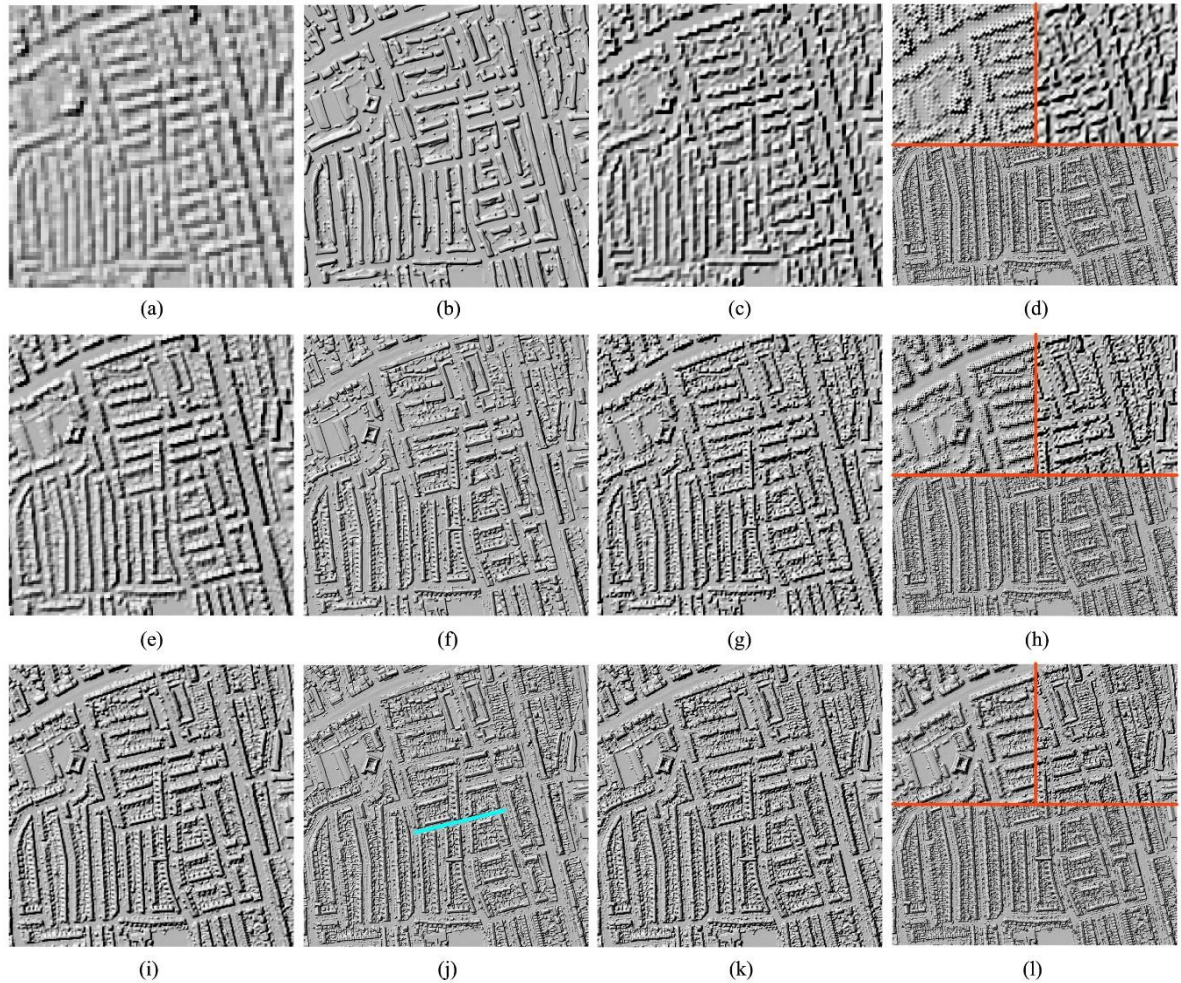

**Fig.5.** Reconstructed results in the study area (zoom-in): (a, e, i) The low-resolution urban DEMs of 8, 4 and 2 m; (b, f, j) The results reconstructed by MSM-CNN using the respective low-resolution DEMs at the same row; (c, g, k) from BI; (d, h, l) Results from IDW (the upper-left part) and CC (the upper-right part), and the reference urban DEM at 0.5 m (the bottom part). The highlight line in (j) is the road centerline for road profiles as showed later in Fig. 9.

The MSM-CNN evidently achieves better results for the reconstructions from all of the three low-resolution urban DEMs (Fig. 5b, f and j). In the whole area, the topographic structure is restored remarkably well, especially for the result reconstructed from the low-resolution

DEM of 8 m, which shows a good fidelity to the actual terrain. The MSM-CNN reconstructed DEM represents well both the continuous and abrupt features. Locally, the buildings and roads are clearly reconstructed, with their boundaries well consistent with the reference terrain. As expected, the restored level of topographic details greatly depends on the input low-resolution urban DEMs and more details are shown in the DEMs reconstructed from input datasets of higher resolutions. In summary, the results indicate that MSM-CNN can effectively achieve the multi-scale reconstruction to enhance the quality of low-resolution urban DEMs.

*4.3. Numerical accuracy*

*4.3.1. Overall accuracy analysis*

Taking the original 0.5 m urban DEM as reference, the results of numerical accuracy assessment of different reconstruction methods are listed in Table 1. From the 2 m low-resolution urban DEM, the 0.5 m product reconstructed by MSM-CNN is the most accurate, confirmed by the lowest MAE (0.194 m) and RMSE (0.918 m); meanwhile, the least accurate reconstruction result is obtained by IDW, which has the highest MAE (0.252 m) and RMSE (1.054 m). The products reconstructed by BI and CC have the same MAE (0.234 m) but different RMSE of 1.012 and 1.028 m, respectively. From the lower-resolution urban DEM of 4 m, the best reconstruction result is still obtained by MSM-CNN, having MAE of 0.316 m and RMSE of 1.295 m. For results from the lowest-resolution dataset of 8 m, the MAE associated with MSM-CNN reconstruction is slightly inferior to that of BI but better than that of CC and IDW; MSM-CNN also returns similar but slightly higher RMSE than the other methods.

**Table 1. Accuracy statistics in the whole reconstructing area**

| Low-resolution urban DEM | Method | MAE (m) | RMSE (m) | STD (m) |
|---|---|---|---|---|

|  | Method | MAE | RMSE | |
|---|---|---|---|---|
| 2 m | MSM-CNN | 0.194 | 0.918 | 0.917 |
| | BI | 0.234 | 1.012 | 1.012 |
| | CC | 0.234 | 1.028 | 1.028 |
| | IDW | 0.252 | 1.054 | 1.054 |
| 4 m | MSM-CNN | 0.316 | 1.295 | 1.290 |
| | BI | 0.328 | 1.325 | 1.325 |
| | CC | 0.332 | 1.357 | 1.357 |
| | IDW | 0.337 | 1.378 | 1.378 |
| 8 m | MSM-CNN | 0.442 | 1.862 | 1.849 |
| | BI | 0.434 | 1.779 | 1.779 |
| | CC | 0.452 | 1.840 | 1.840 |
| | IDW | 0.462 | 1.848 | 1.848 |

Overall, the numerical accuracy of the MSM-CNN reconstructions is mostly higher than that achieved by other interpolation methods. Meanwhile, it is noted that the variances of the numerical accuracy between MSM-CNN and other interpolation methods are not significant, which appears to contrast with the visual analysis of the reconstruction results presented in Fig. 5. The reason may be that the local elevation variation of urban topography in the reconstructing area is relatively small, and the overall statistics may not efficiently reflect the small differences. It is therefore necessary to further investigate this by considering the morphological accuracy for quality analysis as well as conducting numerical accuracy assessment in groups, such as slope ranges and land covers.

*4.3.2. Vertical accuracy based on slope classification*

We further investigate the vertical accuracy based on slope classification. The topographic features are divided into ten ranges according to slope, and then MAE and RMSE are respectively calculated for each of these ranges (Fig. 6). Table 2 lists the average MAEs and RMSEs for all of the ten slope ranges. Herein, the slope data is derived from the original 0.5 m urban DEM. From Fig. 6a-c, a general increasing trend is observed for both MAEs and RMSEs calculated for the different reconstruction results as the slope gradually increases. This indicates that the urban terrain relief as indicated by the slope factor has an obvious influence on the vertical accuracy of DEM reconstruction. As shown in Table 2, among all four approaches, MSM-CNN returns the highest accuracy confirmed by low RMSE and MAE for the reconstructions from all of the adopted low-resolution urban DEMs. The superior accuracy is maintained across all slope ranges until the slope is >= 100%, which covers 76% of the whole reconstruction area.

As the slope of the topography increases to >= 100%, both MAE and RMSE of the MSM-CNN reconstruction results are slightly higher than those of other three methods when the reconstruction is conducted for the low-resolution urban DEM of 8 m. The MAEs of the BI, CC and IDW reconstruction results from the 8 m dataset start to decrease as the slope goes beyond 100%, whereas the their RMSEs continue to increase. In cities, the areas with the slope >= 100% are mostly featured with abrupt change of terrain. Therefore, the reasons for the two aforementioned abnormalities may be because the 8 m low-resolution urban DEM has smoothened out those sharp-fronted topographic features in this area, leading to the disappearance of the abrupt urban topography. As such, the MSM-CNN model may exaggerate the reconstruction error by maximizing the restoration of the abrupt characteristic. For BI, CC, IDW, they essentially smooth the abrupt terrain during reconstruction without recreating abrupt change of the topography. Since the area featured with this highest slope range of >= 100% takes up 24% of the total area, the influence on the reconstruction results is evident. The findings may also explain the overall accuracy assessment result in Table 1, where the MSM-CNN reconstruction result from the 8 m DEM is slightly less accurate than those obtained using other interpolation methods.

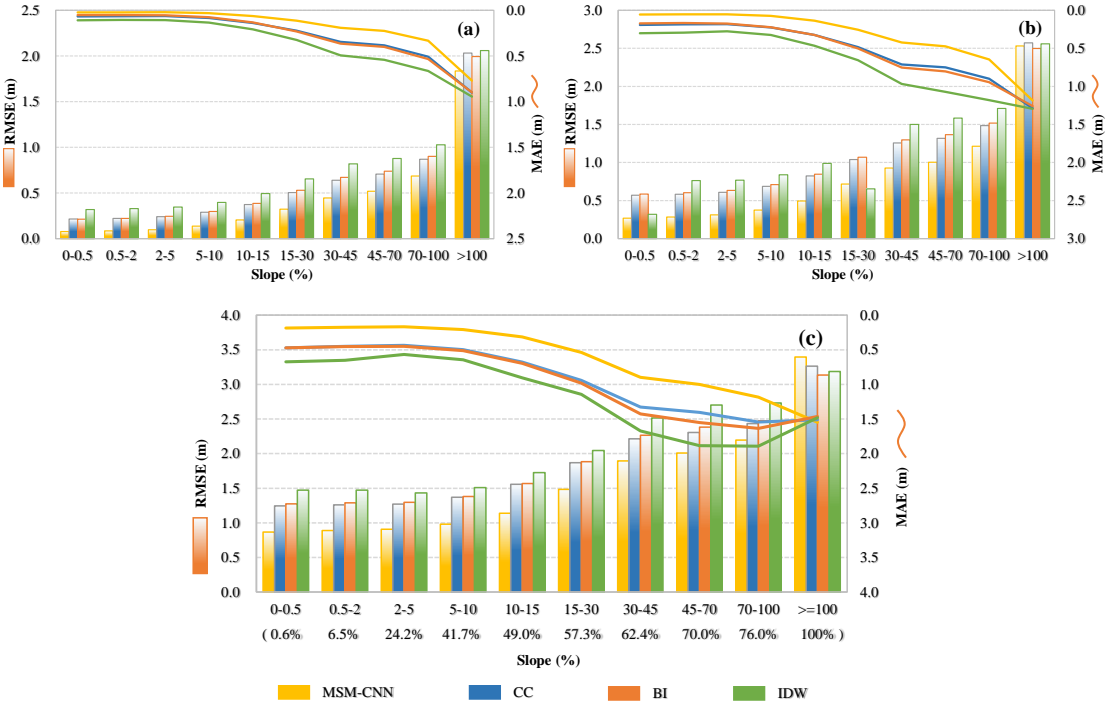

**Fig. 6.** The accuracy statistics for different slope ranges in the whole reconstructing area: (a-c) RMSEs and MAEs calculated for different DEMs reconstructed from the low-resolution 2, 4 and 8 m DEMs. The values inside the bracket below the *x*-axis in (c) are the accumulative frequency of each slope range.

Table 2. Average MAEs and RMSEs calculated for different slope ranges

| Low-resolution urban DEM | Method | Mean of MAE (m) | Mean of RMSE (m) |
|---|---|---|---|
| 2 m | MSM-CNN | 0.179 | 0.441 |
| | BI | 0.279 | 0.620 |
| | CC | 0.278 | 0.609 |
| | IDW | 0.363 | 0.732 |
| 4 m | MSM-CNN | 0.336 | 0.813 |
| | BI | 0.532 | 1.113 |
| | CC | 0.524 | 1.094 |
| | IDW | 0.684 | 1.269 |
| 8 m | MSM-CNN | 0.622 | 1.576 |
| | BI | 0.964 | 1.895 |
| | CC | 0.926 | 1.879 |
| | IDW | 1.152 | 2.079 |

*4.3.3. Vertical accuracy based on land cover classification*

For urban topography, terrain change is closely related to land cover types. Therefore, the vertical accuracy of the reconstructed DEMs from different approaches is also analyzed for various land covers. Herein, the urban land cover is divided into five types for analysis, including roads, buildings, natural environment, multi-surface and other. Natural environment is defined to include those areas representing geographic extent of natural environments and terrain. Multi-surface comprises all of the man-made surfaces that are mainly around buildings, such as yards and plazas. Except for the first four types, the rest is classified as 'other'. Fig. 7 illustrates the distribution of different land covers in a sample area within the case study site.

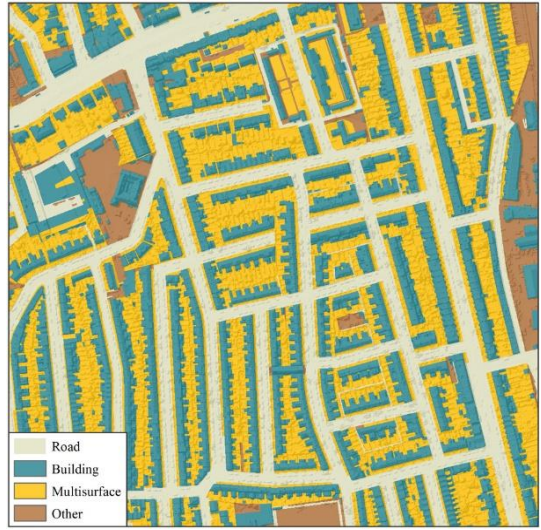

**Fig. 7.** Different types of land covers in a sample area inside the reconstruction site.

Fig. 8 shows the statistics of MAE and RMSE across different land covers for each of the reconstructed urban DEMs. For all of the land cover types, MSM-CNN returns smaller MAEs than all other alternative approaches for all of the reconstruction experiments. But for the 'natural environment' type land cover, the MSM-CNN products reconstructed from 4 m and 8 m low-resolution DEMs have slightly higher RMSE than the results produced by other reconstruction methods. This again demonstrates that the interpolation methods are more suitable for application to natural terrain, but do not produce favorable results for urban topography. In urbanized cities, natural environment is commonly much less dominant than other land cover types, which indicates its influence on the overall accuracy of DEM reconstruction is small. It is interesting to note that, for the road and building land cover types, the MAEs of the MSM-CNN DEMs reconstructed from all 3 low-resolution urban DEMs are much smaller than other reconstructions. Obviously, these are the two major land cover types in urbanized area and cover approximately 40% of the total area in the current study site. The performance analysis results effectively demonstrate that the current MSM-CNN approach offers better capability in restoring urban topographic structure with a high fidelity. In addition, the errors calculated for the multi-surface land cover type are relatively high for all reconstructions although the corresponding topography inherently has a low relief. A possible reason may be that vegetation was not removed from the original 0.5 m urban DEM created from LiDAR data. Vegetation cover may significantly affect the reconstruction accuracy because its elevation changes disorderly and behaves like random noise, which is difficult to be reliably reconstructed from low-resolution urban DEMs.

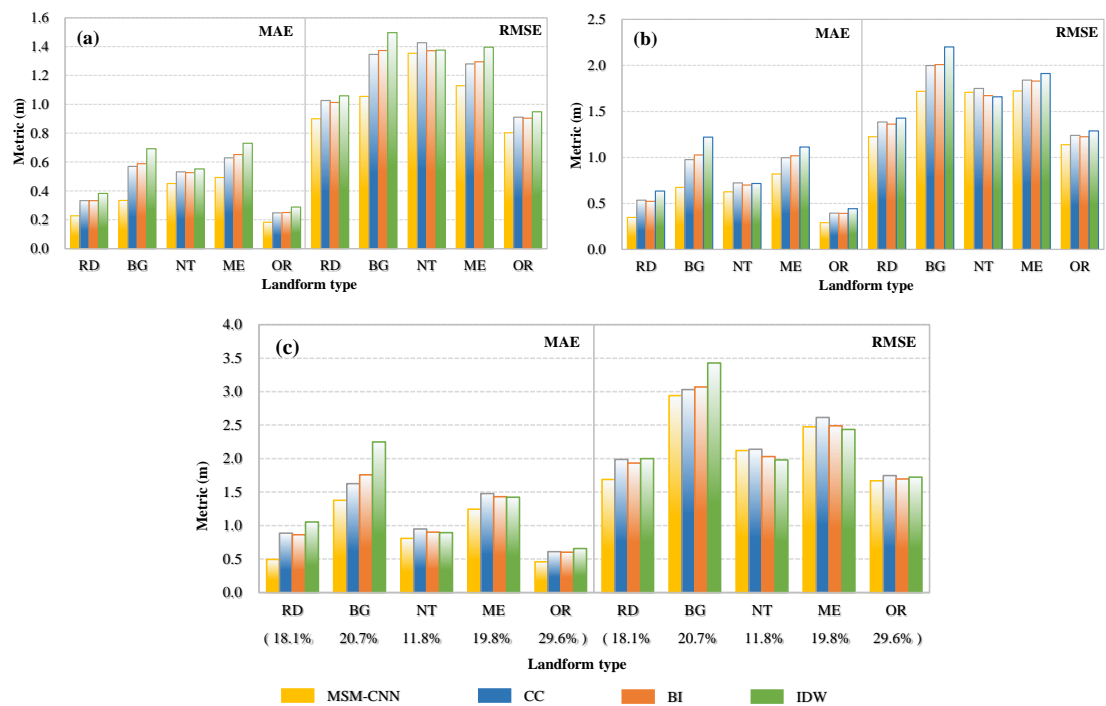

**Fig. 8.** The accuracy metrics calculated for different land covers: (a-c) MAEs and RMSEs for different DEMs reconstructed from the 2, 4 and 8 m urban DEMs. The bracketed numbers below the *x*-axis in (c) indicate the frequency of each land cover: RD-road; BG-building; NT-natural environment; ME-multi-surface; OR-other.

### 4.4. Morphological accuracy

#### 4.4.1. Accuracy assessment based on road profiles

Fig. 9 illustrates the centerline profiles of a road extracted from different reconstructed DEMs. The location of the selected road section is shown in Fig. 5j. Comparing the results obtained using different reconstruction methods, the MSM-CNN road profiles reconstructed from all three lower-resolution urban DEMs show great agreement with the reference profiles extracted from the original 0.5 m dataset. On the contrary, the road profiles generated by BI, CC and IDW show spurious oscillations that are inconsistent with the morphology of urban roads. In particular for the reconstructed results from the lower-resolution 4 m or 8 m urban DEMs, the oscillations in the BI, CC and IDW products are so strong that the centerline profiles can no longer be recognized as a road. Moreover, the three CC road profiles unexpectedly show many deep ditches, which are again inconsistent with normal urban road morphology. The results confirm the much superior capability of the proposed MSM-CNN model in reliably reproducing urban morphology.

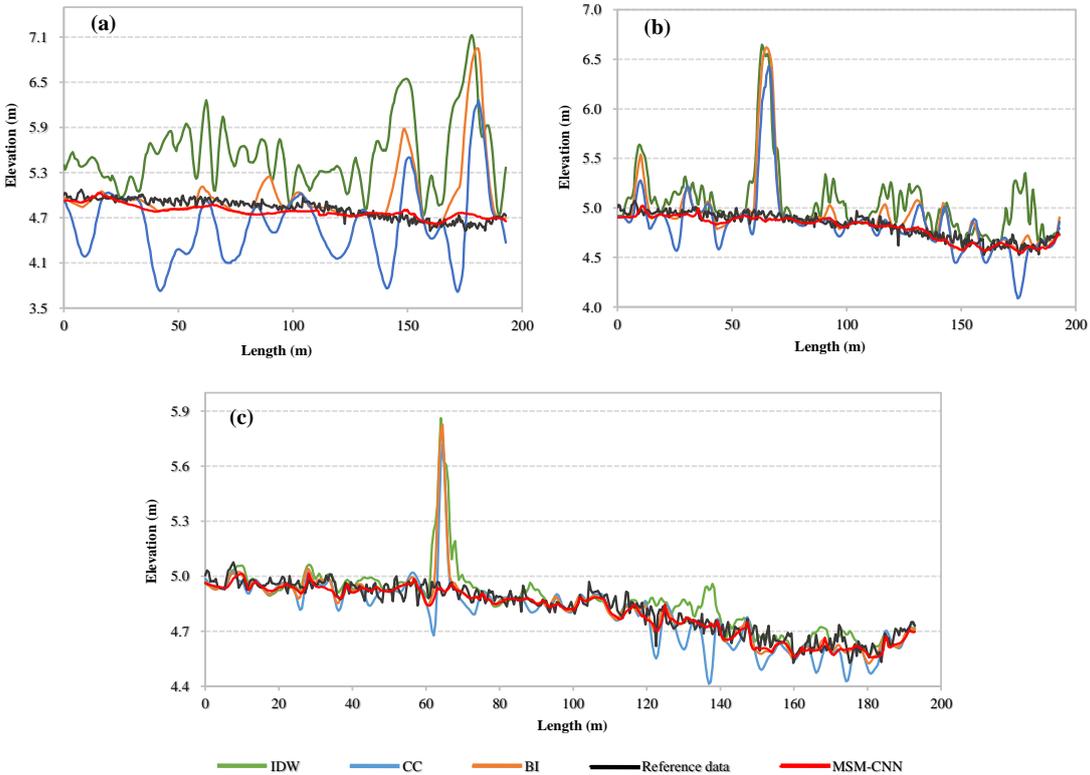

**Fig. 9.** Road profiles extracted from the of Urban DEMs reconstructed using different methods: (a-c) the road profiles from the 8 m, 4 m and 2 m reconstructed urban DEMs, respectively.

Based on the previous accuracy assessment results, BI produces better reconstruction results than the other two interpolation methods. Therefore, the following analysis is focused on comparing the morphological accuracy between the MSM-CNN and BI reconstruction results. Table 3 summaries the statistics of the road-profile variance to quantify the morphological accuracy of the results. For the 4-time reconstructions (i.e., the 0.5 urban DEMs reconstructed from the 2 m equivalent), MSM-CNN clearly gives better result than BI. According to the PCCs calculated for reconstructed road profiles, 51% of the MSM-CNN reconstructed profiles have a PCC greater than 0.95, whereas only 38% of the BI reconstructed profiles reach the same level. For the MSM-CNN and BI reconstructions from the 4 m urban DEM, the difference in the morphological accuracy has significantly increased, as indicated by the average PCC of 0.79 for the MSM-CNN profiles and 0.66 for the BI profiles, respectively. Whilst 51% of the MSM-CNN reconstructed road profiles has the PCC greater than 0.9, only 29% of the BI profiles can reach this level. For 16-time reconstruction, i.e. reconstructing the urban DEMs from 8 m coarse resolution to 0.5 m fine resolution, the improved morphological accuracy achieved by MSM-CNN has become even more prominent, and an improvement of 42% has been achieved comparing with BI. The results demonstrate that the advantage of MSM-CNN in improving the morphological accuracy as represented by road-profile variance becomes more distinct as the resolution of the input urban DEM becomes coarser. In summary, the MSM-CNN reconstruction can substantially enhance the quality of low-resolution urban DEMs through improving morphological accuracy.

Table 3. Morphological accuracy statistics of road-profile variance

| Low-resolution urban DEM | Method | Mean of PCC | STD of PCC |
|---|---|---|---|
| 2 m | MSM-CNN | 0.89 | 0.15 |
|  | BI | 0.83 | 0.20 |
| 4 m | MSM-CNN | 0.79 | 0.24 |
|  | BI | 0.66 | 0.30 |
| 8 m | MSM-CNN | 0.68 | 0.33 |
|  | BI | 0.48 | 0.36 |

*4.4.2. Accuracy evaluation based on building boundary reconstruction*

Using the extraction method described in Subsection 3.2, building boundaries are delineated from the MSM-CNN and BI reconstructed DEMs for comparison, as shown in Fig. 10 in which the reference boundary data is also presented in the vector format. As shown in Fig. 10a for 16-time reconstructions, the overall shapes of the boundaries are reasonably well reproduced by MSM-CNN, although certain fine-level details are smoothened out, which is as expected. However, almost no building boundary can be detected from the BI reconstruction. Fig. 10b illustrates the reconstructions from the 4 m urban DEM, MSM-CNN representation of building boundaries has been further improved and building corners can now be clearly recognized. But BI still fails to reconstruct the overall shape of the building boundaries. As exhibited in Fig. 10c, the building boundaries in the MSM-CNN product reconstructed from the 2 m urban DEM are continuous and close to the reference, while the building boundaries produced by BI are typically segmented and do not align with the reference. Evidently, MSM-CNN outperforms BI in restoring detailed features of urban topography and is more suitable for urban applications.

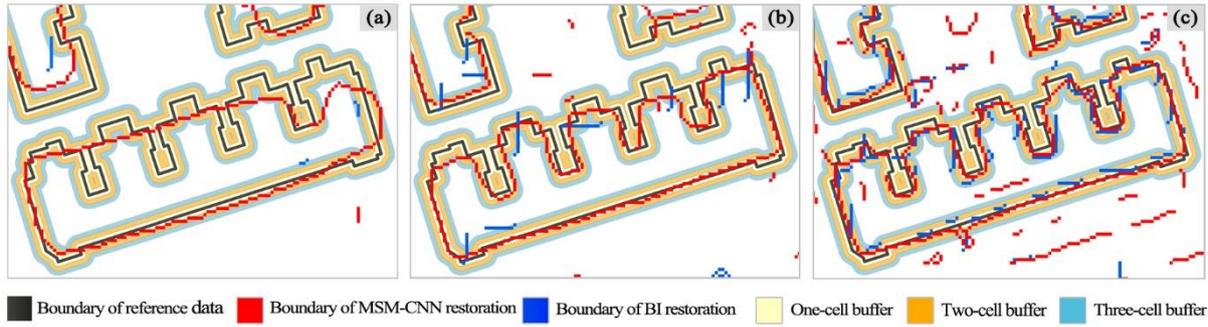

**Fig. 10.** Building boundaries extracted from the reconstructed DEMs: (a-c) DEMs reconstructed from the 8, 4 and 2 m DEMs, respectively. One-cell, two-cell and three-cell buffers are separately the zones with widths of 1, 2, and 3 times the reconstructed cell size around the reference boundary lines.

To quantify the morphological accuracy of building boundary reconstruction, the percentage of correctly restored boundary cells is calculated and plotted in Fig. 11. Overall, comparing with BI, MSM-CNN presents clear superiority, especially for reconstruction from lower-resolution urban DEMs. As expected, regardless the method being used, the morphological accuracy is calculated to be the highest for the 4-time reconstructions for each of the buffer ranges, followed by 8-time and 16-time reconstructions. The accuracy evaluated for the 4-time and 16-time MSM-CNN reconstructions only differs by an average of 2.5 times for the four buffer ranges. However, the accuracy difference unexpectedly reaches 16.2 times for the corresponding BI reconstructions. When the buffer distance is chosen to be three cells (approximately 2 m where the cell size is 0.5 m), the percentage of correctly restored boundary cells returned by MSM-CNN is 70.23% for the 4-time reconstruction, and 34.52% for 16-time reconstruction where the resolution of the input DEM (8 m) is nearly four times larger than the buffer distance. For BI, only 42.73% of the boundary cells are correctly restored by the 4-time reconstruction; for 16-time, the figure has substantially dropped to only 2.91%. This effectively demonstrates that MSM-CNN consistently outperforms BI in restoring building details.

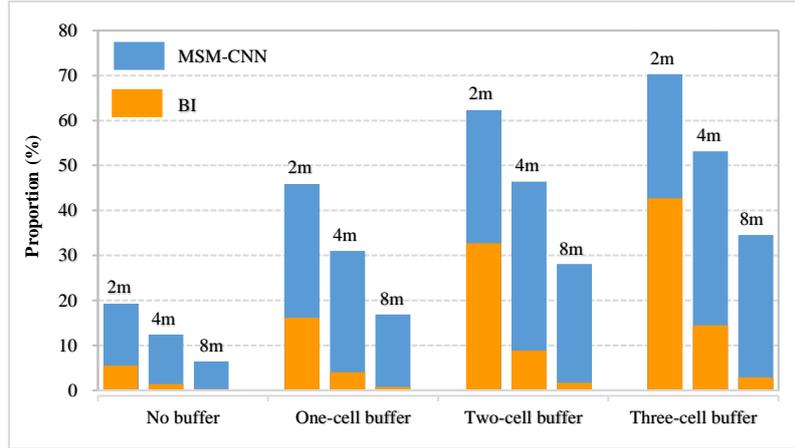

**Fig. 11.** Morphological accuracy statistics of the building boundary. The label of 2, 4 or 8 m on the top of each column denotes that the reconstructed urban DEM is from the 2, 4 or 8 m low-resolution urban DEM.

## 5. Discussion

The proposed MSM-CNN proves that using the deep learning technique for high-resolution urban DEM reconstruction is definitely feasible and has enormous potential for DEM-based applications. In the application of our approach, three points are worth of mention. The first is the range within which it is possible to perform a reconstruction from low to high resolution. It is well known that urban DEMs fail to capture the topographic variations of the complex human-made objects if their resolutions are too coarse. In other words, the high-resolution reconstruction potential from the low-resolution urban DEMs is limited. Hence, according to our empirical knowledge, MSM-CNN perform best if upscaling the resolution from a low-resolution of 10 m and up to 16 times. The second point is the computational costs. The running time of high-resolution reconstruction on a common computing server is approximately one minute for a surface of 50 km$^2$ at a resolution of 0.5 m which indicates that MSM-CNN can be easily applied to large-scale applications. Moreover, recall that we divide the urban DEM of test area into multiple blocks, reconstructing each block and combining the results; this reconstructing process can be speeded up by constructing the blocks in parallel;

further acceleration can be achieved by using more advanced GPU devices. Further improvement to computational efficiency can be achieved considering more network design options. The last point we would like to discuss is the influence of various factors on the reconstruction performance, which will be discussed in detail in subsections 5.1 and 5.2 below.

*5.1. Influence of training data*

It has been widely recognized that the quality of training data has a major influence on the performance of a deep learning model (LeCun et al., 2015; Liu et al., 2017). For MSM-CNN, the reconstruction accuracy is potentially influenced by three factors: 1) typicality; 2) coverage; and 3) scale of the training data. Typicality requires that the training data should represent the typical features of urban topography to be reconstructed. Ideally, the training data should cover typical sample areas of the reconstructing site. For the rapid acquisition of high-accuracy topographic data in these small and typical areas, the unmanned aerial vehicle (UAV) photogrammetry is entirely competent (James and Robson, 2014; Gonçalves and Henriques, 2015; Leitão et al., 2016; Florinsky, 2018). In theory, the larger area the training dataset covers, the more features of the urban topography can be learned. Nevertheless, large coverage of training data inevitably increases the cost of obtaining the sample datasets and training the learning model. Therefore, it is necessary to find a balance between the reconstruction accuracy and the coverage of training datasets. For the implementation of the multi-scale reconstruction approach, this work applies NN downsampling to produce the low-resolution urban DEMs. Although the NN-based downsampled data can validate the MSM-CNN, the effect of different downsampling methods should be further investigated, and it would be better to collect and use real low-resolution datasets for real world applications. Meanwhile, the range between the lower and upper resolutions for training is also better to cover the target range for high-resolution reconstruction.

*5.2. Enhancement with additional terrain information*

The quantitative assessment as presented indicates that the reconstruction accuracy varies with the land covers, slope ranges and details of man-made constructions. This implies that the features of urban DEMs may be better learned by including the additional terrain information to improve reconstruction quality. For example, land covers provide dominant features of urban topography. Land cover types may be considered in the learning process by distinguishing different types of topographic features, e.g. buildings, roads, water surfaces and natural environments (i.e., natural terrain with relatively high relief). With the advanced image classification techniques, the high-resolution remote-sensing imagery is fully capable of mapping the above-mentioned land covers. Terrain attributes, such as slope, curvature or roughness (Wilson and Gallant, 2000), define the multi-dimensional features of urban topography and may be also considered to improve the proposed deep learning process. These attributes can be straightforwardly derived from the corresponding urban DEMs; once the multi-layer attributes are classified, the weight of each layer may be also considered to facilitate a better learning process. Semantic knowledge is another source of information that may be considered. Herein, topographic semanteme refers to the rules of urban construction, for example, the transversal and longitudinal gradients of roads. The semantic knowledge may be utilized to refine the urban topography. Overall, MSM-CNN can be further improved to accommodate more topographic information to further enhance its performance, which deserves attention in future research.

## 6. Conclusions

In this paper, we have proposed an innovative approach, the MSM-CNN, to reconstruct high-resolution urban DEMs from low-resolution equivalents. In order to effectively account for the complexity of urban topography, a multi-scale CNN model is utilized to enhance the reconstruction quality. After the correlations between the low- and high-resolution urban DEMs is learned by the designed training process, the urban DEM at a specified high resolution can be accurately and effortlessly restored from a low-resolution equivalent.

A two-level accuracy assessment procedure including both numerical accuracy and morphological accuracy is also designed to evaluate the performance of the proposed MSM-CNN method, by comparing with other DEM reconstruction methods including IDW, BI and CC. The results show that the high-resolution urban DEMs of 0.5 m can be effectively restored by MSM-CNN from the low-resolution urban DEMs of 2, 4 and 8 m. Also, the MSM-CNN reconstructions are consistently better than the results produced by other methods, in terms of the visual assessment, and also numerical and morphological accuracy.

The promising results presented in this work demonstrates that MSM-CNN has the potential for use in generating high-resolution urban DEMs from low-resolution DEMs, instead of surveying the whole city. In recent years, a number of commercial global DEM products have been released to provide better resolution to represent urban topography, e.g., ALOS AW3D, NEXTMAP World 10, and WorldDEM. These provide rich data source for applying MSM-CNN to reconstruct high-resolution DEMs for cities, which will have profound implications in many applications, including supporting the use of modern flood modelling tools to facilitate more accurate urban flood risk assessment.

**Acknowledgments**

This work was supported by the UK Natural Environment Research Council (NERC) through the WeACT project (grant number NE/S005919/1), ValBGI project (grant number NE/S00288X/1) and Luanhe Living Lab project (grant number NE/S012427/1), the National Natural Science Foundation of China (grant numbers 41501445, 41701450, 41571398), State Major Project of Water Pollution Control and Management (grant number 2017ZX07603-001), China Postdoctoral Science Foundation (grant number 2018M642146), Jiangsu Planned Projects for Postdoctoral Research Funds (grant number 2018K144C), Anhui overseas visiting projects for outstanding young talents in Colleges and universities (grant number gxgwfx2018078), and Key Project of Natural Science Research of Anhui Provincial Department of Education (grant number KJ2017A416).

# References

Abdel-Hamid, O., Mohamed, A.R., Jiang, H., Deng, L., Penn, G., Yu, D., 2014. Convolutional neural networks for speech recognition. IEEE Trans. Audio Speech Language Proc. 22, 1533–1545.

Aguilar, F.J., Agüera, F., Aguilar, M.A., Carvajal, F., 2005. Effects of terrain morphology, sampling density, and interpolation methods on grid DEM accuracy. Photogramm. Eng. Remote Sens. 71, 805–816.

Arun, P.V., 2013. A comparative analysis of different DEM interpolation methods. Egyp. J. Remote Sens. Space Sci. 16, 133–139.

Bagheri, H., Schmitt, M., Zhu, X.X., 2018. Fusion of Urban TanDEM-X raw DEMs using variational models. IEEE J. Sel. Top. Appl. 11, 4761–4774.

Bishop, M.P., James, L.A., Shroder, J.F., Walsh, S.J., 2012. Geospatial technologies and digital geomorphological mapping: Concepts, issues and research. Geomorphology 137, 5–26.

Chen, C., Seff, A., Kornhauser, A., Xiao, J., 2015. Deepdriving: Learning affordance for direct perception in autonomous driving. In: Proceedings of the IEEE International Conference on Computer Vision, pp. 2722–2730.

Chen, Z., Wang, X., Xu, Z., 2016. Convolutional neural network based DEM super resolution. In: International Archives of the Photogrammetry, Remote Sensing and Spatial Information Sciences XLI-B3, pp. 247–250.

Florinsky, I.V., Kurkov, V.M., Bliakharskii, D.P., 2018. Geomorphometry from unmanned aerial surveys. T. GIS 22, 58–81.

Gonçalves, J.A., Henriques, R., 2015. UAV photogrammetry for topographic monitoring of coastal areas. ISPRS J. Photogramm. Remote Sens. 104, 101–111.

Gu, J., Wang, Z.H., Kuen, J., Ma L.Y., Shahroudy, A., Shuai, B., Liu, T., Wang, X.X., Wang, G., Cai, J.F.,

Chen, T.H., 2018. Recent advances in convolutional neural networks. Pattern Recogn. 77, 354–377.

Hawker, L., Bates, P., Neal, J., Rougier, J., 2018. Perspectives on digital elevation model (DEM) simulation for flood modeling in the absence of a high-accuracy Open Access global DEM. Front. Earth Sci. 6, 223.

He, K., Zhang, X., Ren S., Sun J., 2015. Delving Deep into Rectifiers: Surpassing Human-Level Performance on ImageNet Classification. In: Proceedings of the IEEE International Conference on Computer Vision, pp. 1026–1034.He, K., Zhang, X., Ren, S., Sun, J., 2016. Deep residual learning for image recognition. In: IEEE Conference on Computer Vision and Pattern Recognition, pp. 770–778.

Heritage, G.L., Milan, D.J., Large, A.R., Fuller, I.C., 2009. Influence of survey strategy and interpolation model on DEM quality. Geomorphology 112, 334–344.

Hui, Z., Wang, X., Gao, X., 2018. Fast and accurate single image super-resolution via information distillation network. In: IEEE Conference on Computer Vision and Pattern Recognition, pp. 723–731.

James, M.R., Robson, S., 2014. Mitigating systematic error in topographic models derived from UAV and ground-based image networks. Earth Surf. Proc. Land. 39, 1413–1420.

Kingma, D.P., Adam, J.B., 2015. Adam: A method for stochastic optimization. In: International Conference on Learning Representations, pp. 1–15.

Krizhevsky, A., Sutskever, I., Hinton, G.E., 2012. ImageNet classification with deep convolutional neural networks. In: International Conference on Neural Information Processing Systems, pp. 1097–1105.

Lai, W., Huang, J., Ahuja, N., Yang, M., 2017. Deep laplacian pyramid networks for fast and accurate super-resolution. In: IEEE Conference on Computer Vision and Pattern Recognition, pp. 624–632.

Le Besnerais, G., Sanfourche, M., Champagnat, F., 2008. Dense height map estimation from oblique aerial image sequences. Comput. Vis. Image Und. 109, 204–225.

LeCun, Y., Bengio, Y., Hinton, G., 2015. Deep learning. Nature 521, 436–444.

Lee, C.Y., Xie, S., Gallagher, P., Zhang, Z., Tu, Z., 2015. Deeply-supervised nets. In: Artificial intelligence and statistics, pp. 562–570.

Leitão, J.P., de Sousa, L.M., 2018. Towards the optimal fusion of high-resolution Digital Elevation Models for detailed urban flood assessment. J. Hydrol. 561, 651–661.

Leitão, J.P., Moy de Vitry, M., Scheidegger, A., Rieckermann, J., 2016. Assessing the quality of digital elevation models obtained from mini unmanned aerial vehicles for overland flow modelling in urban areas. Hydrol. Earth Syst. Sc. 20, 1637–1653.

Li, J., Wong, D.W.S., 2010. Effects of DEM sources on hydrologic applications. Comput. Environ. Urban 34, 251–261.

Li, X., Shen, H., Feng, R., Li, J., Zhang, L., 2017. DEM generation from contours and a low-resolution DEM. ISPRS J. Photogramm. Remote Sens. 134, 135–147.

Liu, C., Du, W., Tian, X., 2017. Lunar DEM Super-resolution reconstruction via sparse representation. In: 2017 10th Image and Signal Processing, BioMedical Engineering and Informatics, pp. 1–5.

Liu, W., Wang, Z., Liu, X., Zeng, N., Liu, Y., Alsaadi, F.E., 2017. A survey of deep neural network architectures and their applications. Neurocomputing 234, 11–26.

Liu, X., Tang, G.A., Yang, J., Shen, Z., Pan, T., 2015. Simulating evolution of a loess gully head with cellular automata. Chin. Geogr. Sci. 25, 765–774.

Mark, O., Weesakul, S., Apirumanekul, C., Aroonnet, S.B., Djordjević, S., 2004. Potential and limitations of 1D modelling of urban flooding. J. hydrol. 299, 284–299.

Mason, D.C., Trigg, M., Garcia-Pintado, J., Cloke, H.L., Neal, J.C., Bates, P.D., 2016. Improving the TanDEM-X Digital Elevation Model for flood modelling using flood extents from Synthetic Aperture

Radar images. Remote Sens. Environ. 173, 15–28.

Mondal, A., Khare, D., Kundu, S., Mukherjee, S., Mukhopadhyay, A., Mondal, S., 2017. Uncertainty of soil erosion modelling using open source high resolution and aggregated DEMs. Geosci. Front. 8, 425–436.

Moon, S., Choi, H.L., 2016. Super resolution based on deep learning technique for constructing digital elevation model. In: American Institute of Aeronautics and Astronautics SPACE Forum, pp. 1-8.

Moore, I.D., Grayson, R.B., Ladson, A.R., 1991. Digital terrain modelling: a review of hydrological, geomorphological, and biological applications. Hydrol. Process. 5, 3–30.

O'Loughlin, F.E., Paiva, R.C.D., Durand, M., Alsdorf, D.E., Bates, P.D., 2016. A multi-sensor approach towards a global vegetation corrected SRTM DEM product. Remote Sens. Environ. 182, 49–59.

Ozdemir, H., Sampson, C.C., de Almeida, G.A.M., Bates, P.D., 2013. Evaluating scale and roughness effects in urban flood modelling using terrestrial LIDAR data. Hydrol. Earth Syst. Sc. 10, 5903–5942.

Ramirez, J.A., Rajasekar, U., Patel, D.P., Coulthard, T.J., Keiler, M., 2016. Flood modeling can make a difference: Disaster risk-reduction and resilience-building in urban areas. Hydrol. Earth Syst. Sc. Discuss., 1–25.

Rossi, C., Gernhardt, S., 2013. Urban DEM generation, analysis and enhancements using TanDEM-X. ISPRS J. Photogramm. Remote Sens. 85, 120–131.

Szegedy, C., Liu, W., Jia, Y., Sermanet, P., Reed, S., Anguelov, D., Erhan, D., Vanhoucke, V., Rabinovich, A., 2015. Going deeper with convolutions. In: IEEE Conference on Computer Vision and Pattern Recognition, pp. 1-9.

Schmidhuber, J., 2015. Deep learning in neural networks: An overview. Neural networks 61, 85–117.

Shan, J., Aparajithan, S., 2005. Urban DEM generation from raw LiDAR data. Photogramm. Eng. Remote Sens. 71, 217–226.

Simonyan, K., Zisserman, A., 2015. Very deep convolutional networks for large-scale image recognition. In: International Conference on Learning Representations, pp. 1-14.

Tan, M.L., Ramli, H.P., Tam, T.H., 2018. Effect of DEM resolution, source, resampling technique and area threshold on SWAT outputs. Water Resour. Manag. 32, 4591–4606.

Tran, T.A., Raghavan, V., Masumoto, S., Vinayaraj, P., Yonezawa, G., 2014. A geomorphology-based approach for digital elevation model fusion–case study in Danang city, Vietnam. Earth Surf. Dynam. 2, 403–417.

Wilson, J.P., 2012. Digital terrain modeling. Geomorphology 137, 107–121.

Wilson, J.P., Gallant, J.C., 2000. Terrain Analysis: Principles and Applications. Wiley, New York.

Wise, S. 2011. Cross-validation as a means of investigating DEM interpolation error. Comput. Geosci. 37, 978–991.

Xie, S., Tu, Z., 2015. Holistically-nested edge detection. In: Proceedings of the IEEE international conference on computer vision, pp. 1395–1403.

Xu, Z., Wang, X., Chen, Z., Xiong, D., Ding, M., Hou, W., 2015. Nonlocal similarity based DEM super resolution. ISPRS J. Photogramm. Remote Sens. 110, 48–54.

Yue, L., Shen, H., Yuan, Q., Zhang, L., 2015. Fusion of multi-scale DEMs using a regularized super-resolution method. Int. J. Geogr. Inf. Sci. 29, 2095–2120.

Yue, L., Shen, H., Zhang, L., Zheng, X., Zhang, F., Yuan, Q., 2017. High-quality seamless DEM generation blending SRTM-1, ASTER GDEM v2 and ICESat/GLAS observations. ISPRS J. Photogramm. Remote Sens. 123, 20–34.

Zhu, X.X., Baier, G., Lachaise, M., Shi, Y., Adam, F., Bamler, R., 2018. Potential and limits of non-local means InSAR filtering for TanDEM-X high-resolution DEM generation. Remote Sens. Environ. 218, 148–161.